\documentclass[11pt,letterpaper]{article}
\usepackage{cogsys}
\usepackage[T1]{fontenc}
\usepackage{times}
\usepackage[pdftex]{graphicx} 
\usepackage{xspace}
\usepackage{amsmath}

\usepackage{natbib}
\usepackage{comment}
\usepackage{placeins}

\usepackage{array}
\newcounter{rowno}
\cogsysheading{11}{2021}{1-20}{9/2021}{11/2021}

\ShortHeadings{Rational Selection of Goal Operations with Goal-Driven Autonomy}
              {S.\ Kondrakunta, V.\ Gogineni, M.\ Cox, et al.,}

\begin{document} 

\title{The Rational Selection of Goal Operations and the Integration of Search Strategies with Goal-Driven Autonomy}
 
\author{Sravya Kondrakunta}{kondrakunta.2@wright.edu}
\author{Venkatsampath Raja Gogineni}{gogineni.4@wright.edu}
\author{Michael T. Cox}{michael.cox@wright.edu}
\address{Department of Computer Science \& Engineering, Wright State University, Dayton, OH 45431 USA}
\author{Demetris Coleman}{colem404@egr.msu.edu}
\author{Xiaobo Tan}{xbtan@egr.msu.edu}
\address{Dept. of Electrical \& Computer Engr., Michigan State University, East Lansing, MI 48824 USA}
\author{Tony Lin}{tlin@gatech.edu}
\author{Mengxue Hou}{houmengxue@gatech.edu}
\author{Fumin Zhang}{fumin@gatech.edu}
\address{Electrical and Computer Engineering, Georgia Institute of Technology, Atlanta, GA 30332 USA}
\author{Frank McQuarrie}{frankmac@uga.edu}
\author{Catherine R.  Edwards}{catherine.edwards@skio.uga.edu}
\address{Skidaway Institute of Oceanography, Department of Marine Sciences, University of Georgia, Savannah, GA 31411 USA}
\vskip 0.2in
 
\begin{abstract}
Intelligent physical systems as embodied cognitive systems must perform high-level reasoning while concurrently managing an underlying control architecture. The link between cognition and control must manage the problem of converting continuous values from the real world to symbolic representations (and back). To generate effective behaviors, reasoning must include a capacity to replan, acquire and update new information, detect and respond to anomalies, and perform various operations on system goals. But, these processes are not independent and need further exploration. This paper examines an agent's choices when multiple goal operations co-occur and interact, and it establishes a method of choosing between them. We demonstrate the benefits and discuss the trade offs involved with this and show positive results in a dynamic marine search task. 

\end{abstract}

\section{Introduction} 

\emph{Intelligent Physical Systems (IPS)} are cognitive systems that combine perception, actuation, cognition and communication to operate in the physical world. Examples include physical robots, self-driving cars, intelligent homes, and smart grid networks. In general, most IPS need to be capable of robust, long-term autonomy in the presence of uncertain situations, unexpected events, and dynamically changing environments with minimal human intervention. However, achieving such autonomy in a domain independent manner has long been elusive and fraught with difficulty \citep{sidhikhybrid,mota2018incrementally,huntsberger2011intelligent,wilson2013towards,wilson2013domain}. The research presented here proposes an approach that begins to integrate control theory techniques at the lower levels of an IPS with high-level reasoning about system goals. We show the benefit of this approach through a set of marine exploration tasks in complex, underwater environments. 

The cognitive systems community has long acknowledged the importance of agents reasoning about themselves, their behaviors, and their goals\footnote{In this paper, we focus on attainment goals or states to achieve in the external world. Currently, we side-step the issue of applying our approach to performance goals, maintenance goals, or query goals \citep{van2008goals}.}. Several research groups have addressed the issues regarding such full-spectrum reasoning. Approaches that address this include automated planning, belief-desire-intention frameworks, cognitive architecture frameworks, and \emph{goal-driven autonomy (GDA)}. Among these areas, GDA explicitly focuses on the reasoning about goals by implementing various goal operations such as formulation, delegation, and goal change. Goal operations play a vital role in focusing reasoning on the most crucial information, and they also guide the behavior of the IPS when problems arise. 

Here, we extend the idea that goal reasoning comprises a series of primary goal operations that can make an IPS robust and combine these principles with techniques from control theory to provide the foundation for implementing IPS in marine environments. We call this approach goal-driven marine autonomy.
However, since the world is dynamic, instances exist in which the IPS must choose between multiple goal operations that are relevant. The main contribution of this paper lies in developing a procedure for an IPS to make such choices and integrating it into sophisticated problem-solving strategies performed during search applications. We implement the decision processes in a publicly available cognitive architecture named MIDCA. 

The remainder of the paper is as follows. Section \ref{multilayer} describes our multilayer approach to IPS robustness, a framework for building IPS by integrating high-level and low-level autonomy mechanisms. It also introduces the marine problem domain. Section \ref{Search strategies} describes search strategies used by agents to survey the marine domain and presents the results comparing the strategies with and without goal operations. Next, section \ref{MIDCA meta solution} describes an evaluation of a method to improve performance by selecting among relevant goal operations during these strategies. Section \ref{Related research} discusses related research. Finally, Section \ref{Conclusion} presents the closing remarks and future research directions.

\section{A Multilayer Approach to Intelligent Physical Systems} 
\label{multilayer}

\begin{figure}[htbp]
    \centering
    \includegraphics[width=0.85\textwidth]{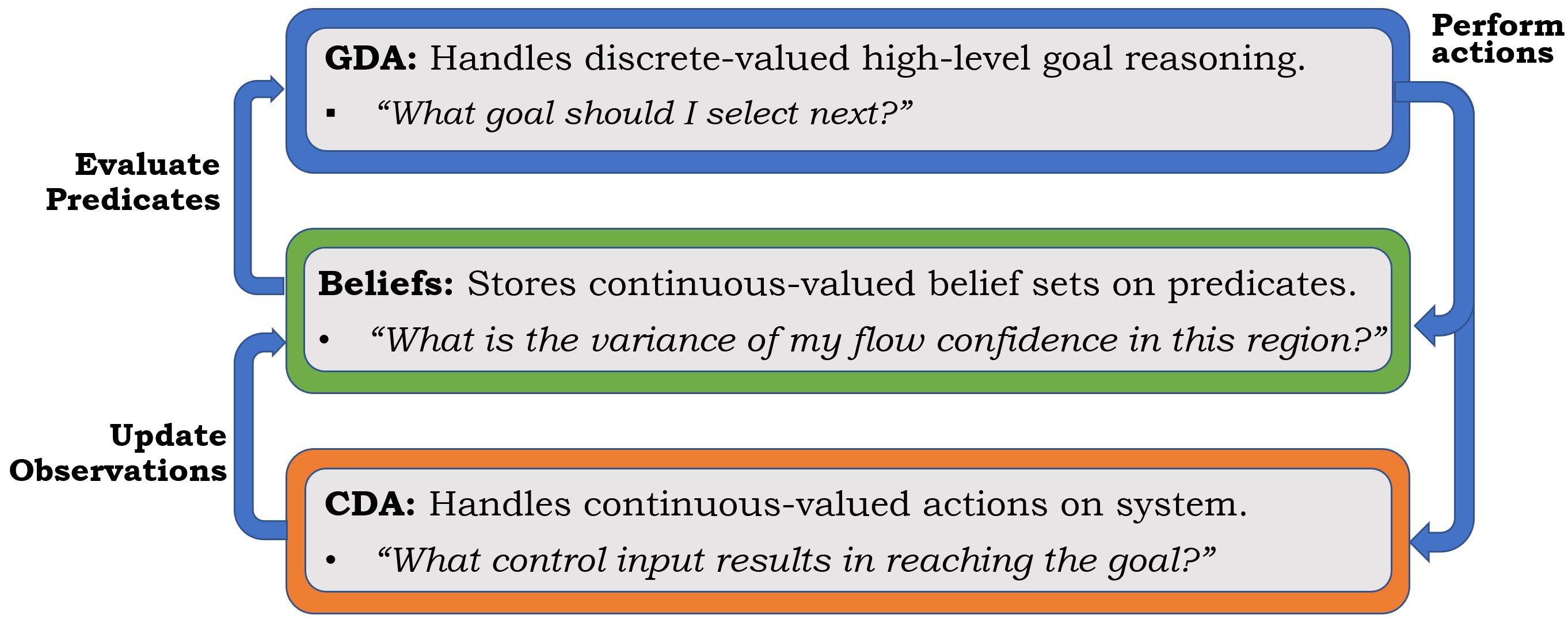}
    \caption{A multilayer approach to intelligent physical systems. At the highest level, GDA performs goal operations given beliefs about the environment. At the lowest level, control manages the actuators and updates continuous sensor values. Mediating between the two, sets of possible states are mapped to discrete predicates and actions are translated to control parameters.}
    \label{fig:multilayer approach fig}
\end{figure}

The objective of our research is to integrate the high-level abstract reasoning of cognitive systems with the preciseness of control theory for complex continuous domains. Such an integration enables the building of robust IPS for applications in dynamic and unpredictable marine environments. We exploit results from a research approach called \textit{goal-driven autonomy (GDA)} that focuses on managing the goals of cognitive systems and merge them with specifics of robotic control theory in what we term \textit{control-driven autonomy (CDA)} (see Figure \ref{fig:multilayer approach fig}). 

With respect to actions, our GDA approach involves goal management and strategies using goal predicate structures and hierarchical task network plans. CDA involves task net optimization and path planning using waypoints and motion commands. With respect to perception, our approach involves problem recognition and state interpretation using problem schemas and explanation patterns.\footnote{Note that we are referring to a kind of internal explanation or self-diagnosis process rather than an external explanation to another agent.} CDA involves sensor fusion and communication generation using vector field models and grid maps. Currently, we have developed the top and bottom layers shown in this figure and await future research to provide the mechanisms of managing uncertainty with belief spaces (but see \cite{lin2020bounded} for preliminary work toward this goal). Furthermore, we have yet to deploy platforms with GDA in field trials. Instead, our AUVs employ CDA. Thus, when we use the term "agent," we are referring to GDA and CDA in simulation. When eventually validated and fielded, the resulting agent aboard an AUV will instantiate our initial IPS.

\subsection{Goal-Driven Autonomy (GDA)} 
\label{GDA}

Goal driven autonomy or GDA \citep{aha2010goal, klenk2013goal, munoz2018adaptive} is a kind of \textit{goal reasoning} \citep{aha2018goal, roberts2018special} applied to issues of robust autonomy in agent-based systems. Unlike standard autonomous systems that generate behaviors given externally provided goals or tasks, the GDA approach is to independently recognize problems that arise, explain what causes the problem, and use the explanation to formulate a goal. Off the shelf planners can then generate sequences of actions that if executed achieve the goal and hence solve the problem. We use the \textit{Metacognitive Integrated Dual-Cycle Architecture (MIDCA)} \citep{cox2016midca} to integrate GDA into the multilayer framework. 

In addition to goal formulation, numerous other \textit{goal operations} exist to manage the goals of the agent. They include goal selection, goal change, goal delegation, and goal monitoring \citep{cox2017goal}. Goal selection selects the goals for the agents to execute. Goal change modifies the agent's goal to a more abstract or specific goal based on resource availability and real-world changes. Goal delegation is helpful in multiagent scenarios to pass goals to other agents when the agent is constrained. Finally, goal monitors surveil the validity of goals based on state changes. In this paper, we focus on goal selection and goal formulation.

\subsection{Control-Driven Autonomy (CDA)}
\label{CDA}
Control theory is the study of dynamical systems control \citep{antsaklis2006linear,khalil2002nonlinear,athans2013optimal}. Its aim is to develop models and algorithms that drive the system to a desired state through system inputs. Control systems vary and are found in systems such as the temperature control in buildings, control of chemical processes in industrial plants, electrical circuits, cruise control in automobiles, and autonomy of robotic platforms and autonomous vehicles. We term control theory applied to autonomous platforms control-driven autonomy or CDA. 

Broadly, there are two categories of controllers: open-loop and closed-loop systems. Open-loop systems control a system purely based on models, whereas closed-loop systems include measurements as feedback to help control the system, usually through correcting behavior. It it is possible to include multiple models for a dynamical system and thus represent multiple modes of operation. Feedback can then be used to select the most appropriate model to accomplish tasks such as fault detection. This mechanism is important in the context of an IPS because of potential damage to the system. For example a drone losing a motor, an underwater glider losing a wing, or a car driving on ice may all require separate models to accurately represent behavior in those situations.

\subsection{The Marine Survey Domain} 
\label{Domain description}

Consider the problem of time-limited surveys of marine environments with autonomous underwater vehicles (AUVs). Typical missions measure salinity, temperature, and pressure throughout the water column and can incorporate acoustic receivers to investigate key aspects of marine life. An key feature within a marine ecosystem is the presence of \textit{hot spots} or regions of high fish density. These areas and the aquatic pathways between them that fish transit represent areas of ecological sensitivity. Thus, discovering the location of major hot spots, especially for endangered species, is an important task. However, many barriers exist in such environments that make mission success difficult. Sea creatures may attach themselves to platforms and slow progress. Tides and currents exist that also impede progress and obstacles may appear requiring course change. Finally, conditions may change, limiting the detection range of acoustic receivers \cite{catherine, Frank_2021}. 

\begin{figure}[htbp]
    \centering
    \includegraphics[width=0.65\textwidth]{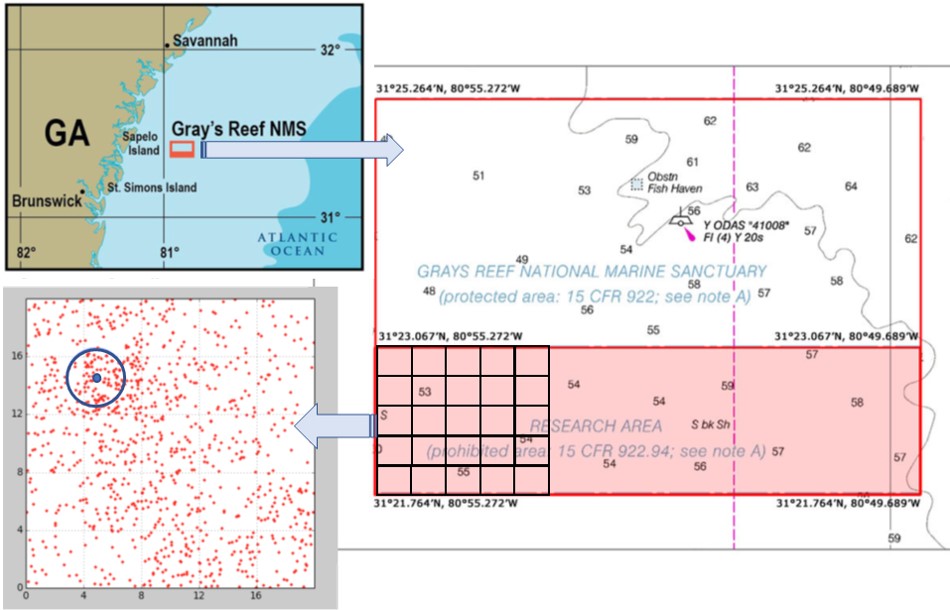}
    \caption{Gray's Reef National Marine Sanctuary is located off the coast of Georgia and contains a research area shown in the insert shaded in pink. Within this, we represent a 5x5 subsection. This grid contains fish hot-spots that are of interest to marine scientists, here within the cell at location (6,14). The agent (indicated by the blue dot) is also in this cell. The circle around the agent indicates the sensor range for detecting acoustic fish tags (the red dots).}
    \label{fig:domain description fig}
\end{figure}

Our research team regularly deploys AUVs such as Slocum gliders and custom robotic fish as part of coastal observing systems, for science-driven experiments and the testing and evaluation of new platforms. During missions, the platforms surface to communicate on regular schedules or in response to forced interrupts. AUV surveys make a valuable contribution to management efforts in Gray's Reef National Marine Sanctuary, located on the inner shelf of the South Atlantic Bight off the coast of Savannah, GA (see Figure \ref{fig:domain description fig}). Gray’s Reef contains fish tagged with transmitters that send an acoustic signal or ‘ping’ at a pre-determined frequency (5 minutes for short experiments, 30-180 minutes for long-term tracking) containing identifiers unique to that instrument, allowing researchers to classify detections by source. 

We implemented a simulator for this domain to test search techniques prior to actual deployment and to empirically evaluate the mechanisms discussed in this paper. The lower right of Figure \ref{fig:domain description fig} shows the portion of the research area modeled by the simulator and split into 25 cells. One cell is shown in the lower left of the figure. The red dots depict 1000 fish (currently assumed to be static) that emit a ping every 17 time steps (time steps are determined to emulate the real-world behavior). In this cell, a hot-spot is located near the co-ordinate (6,14). The blue dot represents a simulated AUV controlled by an agent, and the blue circle represents the receiver detection radius. At Gray's Reef, the detection radius varies with environmental conditions, but currently the simulator assumes it to be 2 co-ordinate units. As mentioned, an agent can identify hot-spots based on the number of pings.

\section{Search Strategies for Aquatic Surveys} 
\label{Search strategies}

An agent's initial goals are to survey the entire region and collect information within a deadline of 600 cycles. If the agent identifies a hot-spot, it needs to report the hot-spot by generating an interrupt, or it can hold the information until its next surface event. The distribution of fish tags is varied to generate successive problems having multiple hot-spots. The agent must find the hot spots and handle all unexpected situations it encounters. 

One unexpected event is biofouling caused by remora attachment. In actual vehicle deployments, remora attachments cause an unexpected change in the vehicle's buoyancy and thus forward speed. The simulator models these effects and injects such anomalies at random. The simulator also allows multiple attachments, thus posing incremental effects. For example, a remora attack reduces the speed by 50\%, and if the agent ignores it, another will further reduce it by 50\%. 

For an agent to survey a region, it must follow a search strategy and handle anomalies that arise concurrently. We integrate three different search strategies. Many search strategies exist in the literature, so implementing, optimizing, and developing the best search strategy is not the focus of this paper. Rather, the strategies below are implementations of existing procedures or minor modifications of them.  

\begin{itemize}
    \item \textit{Structured Search (SS)}: Greedy search or modified hill-climbing search;
    \item \textit{Ergodic Search (ES)}: Control strategy that uses an information metric based on a Gaussian prediction model to guide exploration;
    \item \textit{Ergodic Search Combined with Structured Search (ESCSS)}: Combination of the above two strategies.
\end{itemize} 

SS and ESCSS search each cell in the grid, while ES considers the entire region to be one space and ignores discretization. Subsections \ref{Str search} -- \ref{Sing search} provide details on search strategies. 

\subsection{Structured Search (SS)}
\label{Str search}

Structured search is a modified stochastic hill-climbing algorithm \citep{o1994program}. To overcome the hill-climbing inability to continue when it encounters a local maximum, the agent backtracks and searches further. Table \ref{structured_search_table} presents the procedure for structured search.

The agent starts in an arbitrary cell (line 1), adds it to the visited stack (line 3), searches the current cell for unique pings (line 4) and records all unique pings heard from each of the four boundary edges in a cell. The agent then moves to a neighboring cell in the direction of the highest unique ping density (line 6). Later, it continuously visits the cells with an increasing number of unique pings (lines 9-11). However, when the unique pings in the current cell are less than that of the previously visited cell (line 12), it backtracks and visits the unexplored best neighbor of the last visited cell (line 14). The agent does this iteratively until it visits all cells (lines 7-8) or meets the mission deadline. 

\begin{table}[h]
\setcounter{rowno}{0}
 \centering
 \small
     \caption{Method for structured search (SS)}
     \label{structured_search_table}
        \begin{tabular}{>{\stepcounter{rowno}\therowno.}l}
        \hline
\hspace{0.1 cm}$currentCell \leftarrow startCell$
\hspace{2mm} // \textit{Initial location of the agent}\\
 \hspace{0.1 cm}$visited \leftarrow initStack()$
 \hspace{2mm} // \textit{Keep track of the surveyed cells for backtracking}\\
 \hspace{0.1 cm}$visited.push(currentCell)$\\ 
\hspace{0.1 cm}$currentCellPings \leftarrow \textit{UniquePings}(currentCell)$
 \hspace{0mm} // \textit{Survey current cell and get the no:of fish tags}\\
\hspace{0.1 cm}$loop$ $do$\\
\hspace{0.6 cm}$currentCell, visited \leftarrow \textit{CSearch}(currentCell, visited)$
\hspace{1mm} // \textit{Best neighbor of the current cell}\\
\hspace{0.6 cm}$if$ $visited.empty()$\\
\hspace{1.1 cm}$break$
\hspace{2mm} // \textit{Stop the search as all the cells have been surveyed}\\
\hspace{0.6 cm}$currentCellPings \leftarrow \textit{UniquePings}(currentCell)$\\
\hspace{0.4 cm}$previousCell \leftarrow visited.top()$ 
\hspace{2mm} // \textit{Get previously surveyed cell}\\
\hspace{0.4 cm}$visited.push (currentCell)$
\hspace{2mm} // \textit{Add the current cell to the visited list}\\
\hspace{0.4 cm}$if$ $currentCellPings$ $<$  $\textit{UniquePings}(previousCell)$ \\
\hspace{0.9 cm}$ visited \leftarrow SortByMaxUniqueTags(visited)$\\
\hspace{0.9 cm}$currentCell \leftarrow previousCell$ 
\hspace{2mm} // \textit{Agent is at the top of the hill, perform backtracking}\\
\multicolumn{1}{l}{}\\ 
\multicolumn{1}{l}{\textbf{\textit{CSearch(searchCell, visited)}}}\\ 
    \hspace{0.1 cm}$N, S, E, W = \textit{GetExpectedPings}(searchCell)$
    \hspace{2mm} // \textit{get north, south, east, west expected fish tags}\\
    \hspace{0.1 cm}$SearchCell = \textit{UnvstedMaxCell}(N, S, E, W, visited))$
    \hspace{0mm} // \textit{Unvisited cell with max projected fish tags}\\
    \hspace{0.1 cm}$if$ $searchCell$ $is$ $None$
         \hspace{2mm} // \textit{When the neighbors of the current cell are visited}\\
    \hspace{0.4 cm}$if$ $visited.empty()$\\
    \hspace{0.9 cm}$return$ $searchCell, visited$
     \hspace{2mm} // \textit{Agent has surveyed all the cells}\\
    \hspace{0.4 cm}$else$\\
      \hspace{0.9 cm}$newSearchCell \leftarrow visited.pop()$
    \hspace{1mm} // \textit{Backtrack to the neighbors of the previously visited cells}\\
      \hspace{0.9 cm}$return$ $\textit{CSearch}(newSearchCell, visited)$
          \hspace{2mm} // \textit{Recursively find the best neighbor}\\
    \hspace{0.1 cm}$else$\\
      \hspace{0.4 cm}$return$ $searchCell, visited$
          \hspace{2mm} // \textit{Best neighboring cell with max projected fish tags}\\

        \hline  
        \end{tabular}
\end{table}

\textit{CSearch(searchCell, visited)} is a support function that takes as input a searched cell and the visited cell stack and returns the searched cell's best expected non-visited neighbor (lines 18-20). However, if all the input cell neighbors are visited (line 21), it recursively checks for the best non-visited neighbor from the visited cells and returns it (lines 25-26). 

\subsection{Ergodic Search (ES)}
\label{Erg search}
The Ergodic Exploration for Distributed Information \citep{silverman2013optimal,miller2015ergodic} procedure is a closed-loop solution to the problem of exploration of an unknown space. The search method uses a receding horizon control strategy to optimize the ergodicity metric in \cite{mathew2011metrics}. 

The ergodicity metric relates the time averaged state trajectory $x(\cdot)$ of a dynamic system to a spatial distribution $\phi(x)$. A trajectory is optimally ergodic when the amount of time spent in any given area of a search domain is proportional to the integral of $\phi(x)$ over that same domain. The metric in Equation (1) quantifies the difference between the spatial statistics of $\phi(x)$ and $x(\cdot)$. 
\begin{equation}
\epsilon(x(\cdot)) = \sum^{K}_{k=0}\Delta_k|c_k(x(\cdot))-\phi_k|^2 \label{ergodicMetric}
\end{equation} 
$\phi_k$ and $c_k$ are the Fourier coefficients of the spatial distribution $\phi(x)$ and of the time averaged basis functions along the trajectory $x(\cdot)$. They can be calculated as $\phi_k=\int_x \phi(x)F_k(x)dx$ and $c_k=\frac{1}{T}\int_{t_0}^{t_0 + T}F_k(x(t))dt$, where $F_k(x)=\frac{1}{h_k}\Pi_{i=1}^n\cos({\frac{k\pi}{L_i}x_i})$ are the Fourier basis functions used to approximate the distributions over $n$ dimensions. $h_k$ is a normalizing factor and $x_i$ is the i-th component of $x$.
$K$ determines the number of coefficients used to measure the distance from ergodicity along each of the $n$ dimensions.
$\Delta_k=\frac{1}{(1+||k||^2)^\frac{n+1}{2}}$ is used to place larger weight on lower frequency information. 

In this work, the spatial information distribution $\phi(x)$  is a function of the mean and variance of a gaussian process generated using acoustic tag ping rates over short periods of time. The ergodic metric in (\ref{ergodicMetric}) is then optimized using the algorithm in \cite{miller2015ergodic}. The algorithm works by simulating the control system forward in time over a finite horizon, producing a trajectory and the low level control signal that causes the agent to follow this trajectory. The first value of the control signal is applied and the optimization process is repeated based on the current state of the agent and spatial distribution $\phi(x)$.

\subsection{Ergodic Search Combined with Structured Search (ESCSS)}
\label{Sing search}

As its name suggests, ergodic search combined with structured search combines the two search strategies above. This strategy implements the ES in a single cell, extrapolates the readings obtained at the edges(four directions) in a cell, and performs an SS on the top of ES. In this method, the agent might not always read along all four cell edges because ES is not guided by cell boundaries. The next subsection defines how we build upon the communication established and integrate all three search strategies with goal operations in a cognitive architecture.

\subsection{Goal Operations during Search Strategies}
\label{initial search strategies with cog arch}
As problems arise, the GDA component performs goal operations to solve them. Currently, the system performs two goal operations, although others (e.g., goal change and goal delegation) will be added to the complement in the future.

\begin{itemize}
    \item \textit{Goal Selection}: selects a current goal set among all the goals in the goal agenda
    \item \textit{Goal Formulation}: generates new goal(s) when an agent identifies an opportunity or a problem
\end{itemize}  

\noindent The functioning of the two goal operations is presented formally in detail in Tables \ref{tab:goal_selcect_formalism} and \ref{tab:goal_formulation_formalism}. Table \ref{tab:goal_selcect_formalism} shows the formalism for goal selection \citep{kondrakunta-cox2021}. The goal agenda (${\hat{G}}$) represents all the goals of the agent. Each goal in the goal agenda follows first-order predicate-argument ($p^1(obj1, obj2)$) representation. When given ${\hat{G}}$, the goal selection operation checks for three preconditions ($pre(\delta^{se})$) and results in a single goal or multiple goals for immediate achievement. The first two preconditions ($pre_1(\delta^{se})$, $pre_2(\delta^{se})$) check the validity all goals in ${\hat{G}}$. The third precondition ($pre_3(\delta^{se})$) checks if the resources are sufficient for all goals. Specifically, the $pre_1(\delta^{se})$ checks if the predicates of each goal belong to the class hierarchy tree (${CL}$), which is present in the domain knowledge of the agent. \cite{bergmann2003experience} presents the notation of the ${CL}$, where the are leaf classes that have a root class, and the root class has a superclass. The $pre_2(\delta^{se})$, checks if all the objects of the goal belong to the objects in the domain. Every agent needs to keep track of its own resources to ensure achievement of a maximum number of goals. Hence, the $pre_3(\delta^{se})$, evaluates if the estimated resources from the resource estimator function, $ResourcesForGoals(s,{\hat{G}})$ are sufficient for all goals in ${\hat{G}}$. The resource estimator function returns continuous values. These values are later converted to qualitative values using expert set thresholds for decision making purposes. Finally, After satisfying all the conditions in $pre(\delta^{se})$ the agent can now output a selected goal(s) ($res(\delta^{se})$) by using an appropriate algorithm. For example the paper presents three different algorithms in Sections \ref{Str search}-\ref{Sing search}.  

\begin{table}[htbp]
\centering
\small
     \caption{Formal representation of goal selection}
     \label{tab:goal_selcect_formalism}
        \begin{tabular}{l}
        \hline
        $\delta^{se} ({\hat{G}}: {G}): {G}$ \\
        
         $head(\delta^{se}) = selection$ \\
        
         $parameter(\delta^{se})  = {\hat{G}} = $ $\{p^1(obj1, obj2), p^2(obj3, obj4), ...\}$ \\
         
          $pre_1(\delta^{se}) = p^1 \in {CL} \wedge p^2 \in {CL} \wedge ...$ \\
          
          $pre_2(\delta^{se})  = obj1 \in objs \wedge obj2 \in objs \wedge $
          $obj3 \in objs \wedge obj4 \in objs \wedge...$ \\
          
           $pre_3(\delta^{se})  = ResourcesForGoals(s,{\hat{G}})$ \\
           
           $pre(\delta^{se})  = \{ pre_1(\delta^{se}), pre_2(\delta^{se}), pre_3(\delta^{se}) \}$ \\
           
           $res(\delta^{se}) = g_c = p^x(objy, objz)$ \\
        \hline  
        \end{tabular}
\end{table}

\FloatBarrier Table \ref{tab:goal_formulation_formalism} represents the formalism for goal formulation \citep{kondrakunta-cox2021}. Goal formulation also takes in the goal agenda ${\hat{G}}$, and checks for two preconditions, $pre(\delta^{*})$. The first precondition ($pre_{1}(\delta^{*})$) checks if the agent really needs to formulate a goal and the second precondition checks ($pre_{2}(\delta^{*})$) if the resources of the agent are sufficient to include the new goal. Specifically, the $pre_{1}(\delta^{*})$ checks if the agent observed an anomaly $\omega$ (i.e., an expectation $s_e$ does not match the observation $s_c$) and whether there exist a reasonable explanation $\chi$ for the anomaly. Next, $pre_{2}(\delta^{*})$ generates an estimation for resources of the new goal $ResourceEstimatesFor(g_n)$ and checks if the agent has sufficient resources for the probable goal, $g_n$. If both preconditions are satisfied, the result is a new goal $g_n$ to remove the cause $\neg \omega$ of the anomaly, adding it to the agenda. The resulting goal $g_n$ is generated through explanation patterns as outlined in \cite{gogineni2019probabilistic}.

\begin{table}[htbp]
 \centering
 \small
     \caption{Formal representation of goal formulation}
     \label{tab:goal_formulation_formalism}
        \begin{tabular}{l}
        \hline
        $\delta^{*} ({\hat{G}}: {G): G} $ \\
        
         $head(\delta^{*}) = formulation$ \\
        
         $parameter(\delta^{*})  = {\hat{G}} $ \\
          
          $pre_{1}(\delta^{*})  = \exists\ \chi : \omega \rightarrow (s_e \neq s_c) $\\
          
           $pre_{2}(\delta^{*})  = ResourceEstimatesFor(g_n) $ \\
           
          $pre(\delta^{*})  = \{ pre_1(\delta^{*}), pre_2(\delta^{*})) \}$ \\
           
          $res(\delta^{*}) = g_n= \neg \omega $ \\
          ${\hat{G}} = {\hat{G}} \cup g_n$ \\
        \hline  
        \end{tabular}
\end{table}

In the Marine Survey domain, an agent uses goal selection to select survey goals and goal formulation to formulate goals in the event of a hot-spot or a remora attack. In our implementation, the ES strategy uses only goal formulation, as ES considers the entire grid to be its survey space, and there is only one goal for the agent. This in turn eliminates the possibility for goal selection. In contrast, SS and ESCSS use both selection and formulation. Goal selection chooses the next cell to survey based on the hill-climbing method, and goal formulation generates a goal when the agent perceives a hot-spot or a remora attack.

\subsection{Experimental Design and Results}
\label{Initial results}
We implemented the strategies as three different planning agents (SS, ES and ESCSS). The agents search for hot-spots until reaching a deadline of 600 time units. On their own, these agents do not have any mechanism to cope with anomalies. But the agents can handle anomalies and significantly increase performance when paired with goal operations in MIDCA. MIDCA improves the agents' robustness by detecting anomalies during plan execution and dynamically formulating new goals to handle them at run-time. For example, remora attachments prompt goals to clear the remora from the vehicle. Control behavior that achieves such goals include flying backward.

We created three hot-spot patterns or scenarios in the 5x5 survey region: the first contains four hot-spots at the outer corners of the region, the second contains four hot-spots at the inner corners of the region, and the third contains five hot-spots, where four are at the inner corners, and one is at the center of the region. A trial requires an agent to traverse the 25 cells of a scenario starting from a given initial location and find all hot-spots before the deadline. With 100 initial locations for each scenario (each initial location is a specific real-world latitude and longitude co-ordinate, and the co-ordinate could be in any part in each cell). Since three scenarios, an agent performs 300 trials. The output of one trial in this search will yield 25 cells to classified as either hot-spot or not. Therefore, 300 trials will yield 300x25=7500 data points to derive F1 scores\footnote{F1 is the harmonic mean of precision and recall. Here precision is the fraction of true positives among the instances classified as hot-spots, while recall is the fraction of classified instances among all actual hot-spots.} for each search strategy. 

Table \ref{Search Strategies with Anomalies} presents results for the strategies with and without anomaly responses, i.e., with just goal selection and with both goal selection and goal formulation. When agents ignore anomalies and perform only goal selection, the ES strategy performs best. That is, a raw control theory technique without goal reasoning proves most successful when ignoring remora attacks. However, all strategies improve when combined with high-level reasoning that handles such situations. But out of the three, ES performed the worst on average, while ESCSS and SS performed equally well with a marginal edge to the former. 

\begin{table}[htbp]
\centering
\small
\caption{Search strategies with anomaly response}
\label{Search Strategies with Anomalies}
\begin{tabular} {| p{2.8 cm} | p{1.5 cm} | p{1.0 cm} | p{1.0 cm} | p{1.0 cm} |}
  \hline 
   \textbf{Goal Operations}&\textbf{Parameter} &\textbf{SS}	&\textbf{ES} &\textbf{ESCSS}\\
     \hline
  &Accuracy	&0.830 &0.839 &0.834\\
 With only  &Recall     &0.140 &0.192 &0.152\\
 selection &Precision	&0.538 &0.613 &0.577\\
  &\textbf{F1 score}	&\textbf{0.223} &\textbf{0.293} &\textbf{0.241}\\
  \hline
   &Accuracy	&0.910 &0.875 &0.916\\
   With &Recall     &0.527 &0.373 &0.566\\
   selection and &Precision	&0.923 &0.795 &0.925\\
  formulation &\textbf{F1 score}	&\textbf{0.671}  &\textbf{0.508} &\textbf{0.702}\\
  \hline
\end{tabular}
\end{table}

Despite these averages, each strategy has its advantages under certain conditions. For example, SS performs better in the cells located at corners than ES or ESCSS, whereas ES is better in the grid center cells. We claim that such insights can benefit the agent as a heuristic method, and we will demonstrate this in the next section. Yet as mentioned previously, optimizing  search is not the focus of this paper. Instead, we provide such heuristics as domain-dependent rules and demonstrate the trade offs involved. 


Even with such a capability, the performance improvement obtained would be specific to only one type of anomaly (remora). More realistically, multiple anomalies types arise in real-world settings, and sometimes responding to some anomalies is just a strain on resources. For example at the end of a mission, it is more productive to perform goal selection (i.e., survey another cell) than to respond to a remora attack given the limitation of time. Although slower, the platform may find another hotspot; whereas, by spending the effort to remove the remora, the deadline may expire. Therefore, we provide some domain-independent rules to handle such situations and further improve the agents' robustness in the case of unknown anomalies. Furthermore, these general rules also form the basis for choosing between its goal operations. The following section provides the technical details and an evaluation of the result.

\section{Agents that Select Goal Operations (ASGO)}
\label{MIDCA meta solution}

To further improve the performance of the agent, we provide the agent with both domain-specific and domain-independent generic rules. The domain-specific rules help the agent improve each of the individual search strategies; the domain-independent rules help the agent be flexible in choosing the right goal operation at the right time. To test the performance of these new rules, we introduce two different kinds of anomalies that exist in the real world. One is the flow in water and blockades; flow displaces the agent from its current location, and blockades hinder the agent's movement from one location to another. Whenever an unknown anomaly occurs, the agent might not be in a position to pursue its current goal. In that case, the agent has two options: Select another goal from the remaining set of goals or formulate a goal to respond to the anomaly. These responses include a knowledge goal to investigate the new anomaly or make the agent enter a safe control mode to protect itself from the unknown anomaly. \textit{Agents that Select Goal Operations (ASGO)} can choose between these two operations, goal re-selection (which is essentially goal selection) and goal formulation, smartly.

\begin{table}[h]
\setcounter{rowno}{0}
 \centering
 \small
     \caption{Method for selecting goal operations. 
     Parameter $\Sigma$ is the domain knowledge of the agent, $s_{c}$ is the observed state of the agent, $s_{e}$ is the expectations of the agent, $g_c$ is its current goal, and $\hat{G}$ is its goal agenda.}
     \label{tab:midca_meta_response}
     \begin{tabular}{>{\stepcounter{rowno}\therowno.}l}
    \hline
    \multicolumn{1}{l}{\textbf{\textit{ExecGoalOperations ($\Sigma,s_c,s_e,g_c,\hat{G}$)}}}\\
    \hspace{0.5 cm}$ \pi \leftarrow \Pi(\Sigma, s_c, g_c) $     \hspace{2mm} // \textit{Plan to achieve the current goal}\\ 
    \hspace{0.5 cm}$while$ $R(s_c) > 0$  $do $ \hspace{2mm} // \textit{Loop until there are enough resources}\\ 
    \hspace{1 cm}$ s_c \leftarrow \gamma(s_c, \pi[1]) $ 
    \hspace{2mm} // \textit{Execute the first action and add its results to the agent's current state}\\ 
    \hspace{1 cm}$s_e \leftarrow s_e \cup pre(\pi[1]) \cup \pi[1]^+ - \pi[1]^- $
    \hspace{2mm} // \textit{Agent's expectations from the results of the first action}\\
    \hspace{1 cm}$\pi \leftarrow \langle \alpha_2, \alpha_3, ... \alpha_n \rangle $
    \hspace{2mm} // \textit{Remaining Plan}\\
    \hspace{1 cm}$if$ $s_c \not\models s_e$ $then$     
    \hspace{2mm} // \textit{When Agent's expectations donot match its observations (anomaly)}\\
    \hspace{1.5 cm}$g_f \leftarrow \beta(s_c, g_c)$
        \hspace{2mm} // \textit{Agent formulates a new goal}\\
    \hspace{1.5 cm}$g_s \leftarrow \delta^{s_e} (s_c, \hat{G})$
     \hspace{2mm} // \textit{Agent selects a new goal}\\
    \hspace{1.5 cm}$g_{aff} \leftarrow \textit{AllGoalsAffected}(s_c, s_e, \hat{G})$  \hspace{2mm} // \textit{Agent estimates goals effected by the anomaly}\\
      \hspace{1.3 cm}$if$ $g_s$ $in$ $g_{aff}$ $then$
       \hspace{2mm} // \textit{When the agent's selected goal is effected }\\
      \hspace{2 cm}$g_c \leftarrow g_f$ 
      \hspace{2mm} // \textit{Agent prioritizes the formulated goal}\\
      \hspace{2 cm}$\hat{G} \leftarrow \hat{G} \cup g_c$
        \hspace{2mm} // \textit{Agent updates its goal agenda}\\
    \hspace{1.3 cm}$else$\\ 
    \hspace{2 cm}$g_c \leftarrow g_s$  \hspace{2mm} // \textit{Agent prioritizes the selected goal}\\
      \hspace{1.3 cm}$ \pi \leftarrow \Pi(\Sigma, s_c, g_c) $  \hspace{2mm} // \textit{Plan to achieve the prioritized goal}\\
    \hspace{0.8 cm}$if$ $\pi = \langle \rangle$ then
    \hspace{2mm} // \textit{When the plan is empty}\\
    \hspace{1.3 cm}$\hat{G} \leftarrow \hat{G} - g_c$
        \hspace{2mm} // \textit{Agent removes the current goal from its goal agenda}\\
    \hspace{1.3 cm}$g_c \leftarrow \delta^{s_e} (s_c, \hat{G})$
      \hspace{2mm} // \textit{Performs goal selection}\\
    \hspace{1.3 cm}$ \pi \leftarrow \Pi(\Sigma, s_c, g_c) $ 
    \hspace{2mm} // \textit{Plans to achieve the selected goal}\\
        \hline
        \end{tabular}
\end{table}

Table \ref{tab:midca_meta_response} represents ASGO's method for selecting a goal operation among multiple competing ones. Following the notation in \cite{ghallab2004automated}, the procedure takes the following inputs: Environment model $(\Sigma)$, the current observations of the world $(s_c)$, the expected observations of the world $(s_e)$, current goal $(g_c)$ and a goal agenda $(\hat{G}$ = $\{g_1, g_2, ... g_c, ... g_m\})$. ASGO creates a plan $(\pi = \langle \alpha_1, \alpha_2, ... \alpha_n \rangle)$ of $n$ actions to achieve its initial goal $(g_c)$ (line 1) and starts executing each action in the plan. While doing so, it perceives observations $(s_c)$ from the world (line 3) and obtains expected observations $(s_e)$ from the results of the action that is currently being executed (line 5). Later, when it executes all the actions in the plan, ASGO removes the current goal from its goal agenda $(\hat{G})$ (line 17), applies goal selection operation to select a goal from its goal agenda (line 18), and plans to pursue the goal. ASGO follows the process mentioned above until it runs out of resources $(R(s_c))$ (line 2) or when an anomaly occurs. An anomaly is a condition where the observed state significantly differs from the expected state (line 6)  \citep{dannenhauer2016informed}. When ASGO runs out of resources, it halts. However, when an anomaly occurs and hinders its ability to achieve its current goal, it reasons about the anomaly to formulate a goal ($g_f$) as a response (line 7). Similarly, since the current goal is no longer viable to pursue, ASGO performs goal selection to pursue a different goal ($g_s$) from its goal agenda (line 8). 

To determine which goal to pursue between the formulated goal $g_f$ and selected goal $g_s$, ASGO creates an estimate of all the goals affected ($g_{aff}$) by the anomaly (line 9) and pursues the formulated goal $g_f$ when the selected goal $g_s$ is in the affected goals (lines 10-11). Otherwise, it chooses the selected goal $g_s$ to pursue (line 14). Some limitations of the ASGO agent include: as we add more goal operations, the rules will increase, hence taking a longer retrieval duration.

\subsection{Empirical Results with Domain-Specific Rules}
\label{Main results_domain_specific}

Before implementing the ASGO agent, Let us look at an agent that improves each of the individual goal operations mentioned in subsection \ref{Initial results} with domain specific rules from the experiments on individual search strategies. The rules are tailored to the domain and are provided by a human expert with the data observations from the previous experiment. The rules are generally tied to the location of the agent in the domain and the resource availability. The major rules include:
\begin{itemize}
    \item If the agent is in the corner cell and has sufficient resources to achieve all its goals, it should perform goal selection using SS;
    \item If the agent is in the center cell and has sufficient resources to achieve all its goals, it should perform goal selection using ESCSS;
    \item If the agent is in any cell and does not have sufficient resources to achieve its goals, it should abandon all the remaining goals and selects ES on the whole search area;

\end{itemize} 
These rules aid both goal operations. Let the new agent using these rules be called \textit{Agent with Improved Goal Operations (AIGO)}.  

Table \ref{Search Strategies with operations} compares the data from four different agents: The first three agents are the same agents mentioned in the previous subsection \ref{Initial results}, the fourth agent is the AIGO. The data collected is over the pattern presented in Figure \ref{fig:domain description fig} with a single hot-spot. The data is collected over 100 different trials by varying the agent's initial location; this experiment provides 2500 data points to calculate the F1 scores. The anomalies in this test case are once again the remora attacks.

\begin{table}[htbp]
\centering
\small
\caption{Search Strategies with Goal Operations}
\label{Search Strategies with operations}
\begin{tabular} {| p{2 cm} | p{1 cm} | p{1 cm} | p{1 cm} | p{1 cm} |}
  \hline 
   \textbf{Parameter} &\textbf{SS}	&\textbf{ES} &\textbf{ESCSS} &\textbf{AIGO}\\
    \hline 
Accuracy	&0.968 &0.937 &0.966 &0.998\\
Recall     &0.990 &1.000 &1.000 &0.960\\
Precision	&0.559 &0.390 &0.543 &1.000\\
\textbf{F1 score}	&\textbf{0.715}  &\textbf{0.562} &\textbf{0.704} &\textbf{0.979}\\
  \hline
\end{tabular}
\end{table}

\FloatBarrier As the data depicts, agents SS, ES, and ESCSS output a similar pattern of results as in Table \ref{Search Strategies with Anomalies}. ES performed the worst, and both SS and ESCSS performed comparably. The method that outperforms all individual strategies is AIGO, which uses generic rules to improve individual goal operations. If the rules always hold, then the agent's performance is near-perfect. Since the world is dynamic, the rules might not always hold; even if the rules do hold, the agent might not implement them due to unforeseen reasons. In such situations, multiple goal operations co-occur, and the agent must intelligently choose a goal operation to improve performance. Next, we elaborate on responding to such problems by implementing the ASGO agent.

\subsection{A Working Example of ASGO in the Marine Survey Domain}
\label{example}

An example scenario from the Marine Survey Domain can help us understand the prioritization of goal operations by the ASGO agent. Moreover, this will also present the importance of such prioritization for an agent in the real world.

\begin{figure}[htbp]
    \centering
    \includegraphics[width=0.4\textwidth]{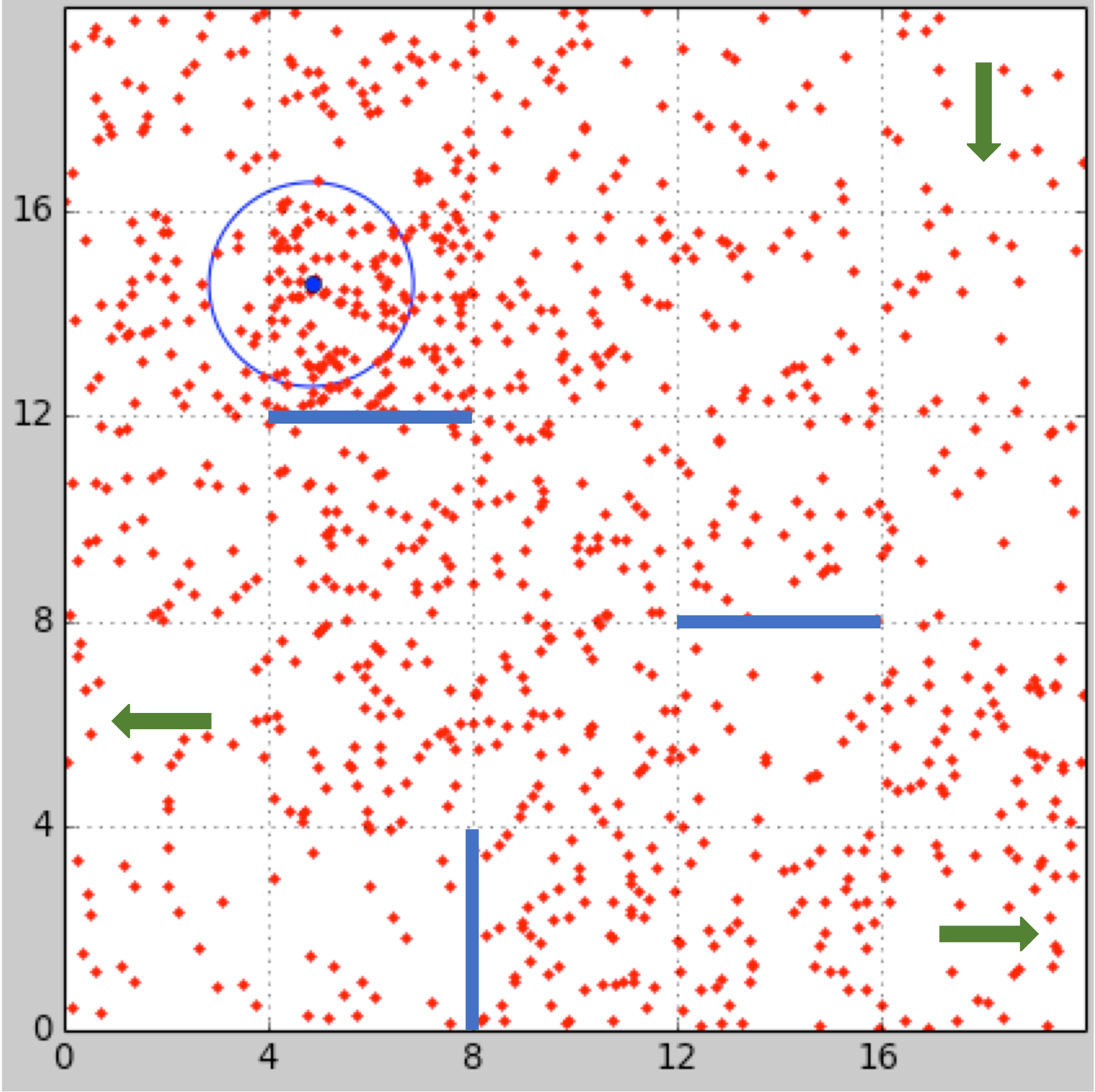}
    \caption{An example in the Marine Survey Domain with Remora attacks and Blockades (indicated by the black Lines). The agent (indicated by the blue dot) is at the location (6,14). The circle around the agent indicates the sensor range for detecting acoustic fish tags (the red dots).}
    \label{fig:example fig}
\end{figure}

Figure \ref{fig:example fig} represents an example scenario in the Marine Survey Domain. As mentioned, there are twenty-five survey goals for the ASGO agent. To test the performance of the ASGO agent, we test the agents' performance against several types of anomalies. First, Remora attacks hinder the agent's movement; second, Blockades (represented in blue lines) hinder the agent's movement from one location to the blocked location. Finally, flow (represented in green arrows) displaces the agent in the direction of the flow.

Consider a scenario where the ASGO agent performs goal selection and surveys the location (6,14). It then encounters a Remora attack. The agent has two choices; select a new survey location (Goal Selection), or formulate a goal to glide backward to respond to the Remora attack (Goal Formulation). Since the Remora attack hinders the agent's movement, affecting all the survey goals, including the goal of selecting any new survey location. So,  the ASGO agent, as per the algorithm \ref{tab:midca_meta_response}, prioritizes goal formulation and glides backward to free itself from the Remora attack; and completes the survey in (6,14). After which, it selects the survey goal (6,10). So, the agent must move from its current location (6,14) to the destination location (6,10).

The ASGO agent then encounters a blockade anomaly that hinders movement from (6,14) to (6,10). Thus, it now has two choices; select a new survey location (10,14) (Goal Selection) or formulate a new goal to inspect the entrance to understand the cause of blockade. In this scenario, the blockade does not affect the goal to survey the location (10,14). So, the agent prioritizes goal selection and surveys location (10,14). 

Thus, ASGO prioritizes goal formulation and goal selection operation based on the goals affected by the anomaly. If the selected goal is affected, the agent prioritizes goal formulation or else goal selection. Such prioritization helps the agent address its goals promptly.

\subsection{Empirical Results with Domain-Independent Rules}
\label{Main results}

We consider two additional anomalies in this test case, namely blockade and flow. Having a single hotspot, the fish distribution scenario for these results is the same as shown in Figure \ref{fig:domain description fig}. We use the same 100 initial locations as in Section \ref{Initial results} and thus obtain 2500 data points. As mentioned, we introduce a new agent that uses domain-specific rules to enhance its performance. The new agent utilizes domain-specific rules to select a best-fitting search strategy corresponding to its location and resources. This new agent has a performance increase compared to the agents that use a single search strategy (SS, ES, ESCSS). We have two versions of such an agent: \textit{Select First} agent, which always gives priority to goal selection, and the \textit{Formulate First} agent always prioritizes goal formulation. We compare their performance against that of the ASGO strategy.

The first set of results in Table \ref{Intelligent iteraction of operations} compares agent performance when flow anomaly occurs with the existing remora attacks. The second set includes one other kind of anomaly, blockade. The algorithm is generalizable to any other future anomalies. As the values depict, both Select-1st and Formulate-1st agents perform similarly, but ASGO outperforms them.

\begin{table}[htbp]
\centering
\small
\caption{Agent performance with intelligent decision making between goal operations}
\label{Intelligent iteraction of operations}
\begin{tabular} {| p{1.5 cm} | p{1.5 cm} | c | c | c |}
  \hline 
   \textbf{Anomalies} &\textbf{Parameter}	&\textbf{Select-1st} &\textbf{Formulate-1st} &\textbf{ASGO}\\
   \hline 
Remora &Accuracy	&0.984 &0.983 &0.993\\
and &Recall      &0.620 &0.670 &0.840\\
flow &Precision	 &0.984 &0.893 &0.988\\
&\textbf{F1 score}	&\textbf{0.760} &\textbf{0.766} &\textbf{0.908}\\
    \hline 
Remora, &Accuracy	 &0.981 &0.984 &0.990\\
block &Recall      &0.540 &0.650 &0.790\\
and &Precision	 &0.981 &0.928 &0.975\\
flow &\textbf{F1 score}  &\textbf{0.697} &\textbf{0.765} &\textbf{0.872}\\
  \hline
\end{tabular}
\end{table}

\section{Related research} 
\label{Related research}

 Similar research work that incorporates both control and cognitive level intelligence includes, the \textit{Control Architecture for Robotic Agent Command and Sensing (CARACaS)}, developed for autonomy in underwater platforms \citep{huntsberger2011intelligent}. CARACaS uses intelligent decision-making to make the unmanned underwater agent robust. CARACaS uses three different components to make such decisions. The first is a behavioral engine which is the core of the system. It is used in real-time for several generic behaviors of the agent. Second is a perception engine which produces maps. Agents use these maps for navigation purposes. Finally, there is a dynamic planner called Continuous Activity Scheduling Planning Execution and Re-planning which is used to handle goal-based planning and re-planning operations on-board. In this work, the burden is on the planner for goal prioritization and goal achievement. However, in our work we use planner for just goal achievement and perform goal selection operation for choosing goals.
 
 Another research effort in underwater platforms that also uses goal reasoning is \cite{wilson2013towards,wilson2013domain}. The work formulates goals in unexpected situations using social, opportunity and exploration motivators. It then re-prioritizes the agent's goals based on the new goals. It also presents an approach to integrating high-level intelligent behavior with motion autonomy while extending the work in multiagent scenarios. In addition more recent works from \cite{wilson2018goal} presents a goal manager which chooses goals based on a priority value. The goal formulator assigns a priority value when it generates a goal and provides it to the goal manager. Although Wilson shares common interests with the current paper, they do not address the issue of interactions between multiple goal operations. Also, \cite{schoenecker2018goal} uses goal reasoning to improve a sonar sensor's performance, but avoids the interactions. In addition, \cite{niemueller2019goal} uses rule based approach to integrate execution and planning. It uses goal life cycle model \citep{roberts2014goal} to handle goal-mode transitions. However, it does not address the decision making involved when multiple goal-transitions occur.

Goal-driven autonomy and goal reasoning more generally are relatively new frameworks used in research compared to many technical areas of AI, but they are active in the cognitive systems community. The applications of goal reasoning include its usage in the \textit{Autonomous Response to Unexpected Events (ARTUE)} system \citep{klenk2013goal}. \cite{shivashankar2014towards} outlines solutions to a number of goal reasoning challenges. This work uses a hierarchical goal network structure to decompose a higher-level goal into several sub-goals to overcome particular limitations using \textit{Motivated ARTUE (M-ARTUE)}. ARTUE and MIDCA agents both employ goal reasoning to identify and respond to unexpected situations in a dynamic environment, both use goal reasoning to perform several operations on their goals. Some of the works mentioned here might not use goal reasoning explicitly, but they share similar goal-based behaviors. \cite{roberts2014goal, roberts2021goal} presents work on goal-life cycle which closely ties with the idea of goal operations. The work describes phases of a goal from its initial formulation to achievement. It does not address the interaction between multiple goal operations. However, more specifically \cite{gogineni2019probabilistic} presents goal formulation through explanation patterns and anomaly detection. Goal formulation based on domain-independent heuristics called motivators (opportunity, exploration, and social), where each motivator is weighed based on urgency and fitness, are presented in M-ARTUE \citep{wilson2013domain}. On similar terms, \cite{karneeb2018distributed} presents discrepancy detection in air combat agents. The paper focuses on discrepancy detection and response in the real-world. However, none of the works mentioned explicitly dive into the decision making process when multiple kinds of goal operation co-occur.

One reason for the applicability of goal reasoning in such diverse environments is due to its ability to perform various operations on its goals. For example, \cite{rabideau2009tractable} (inspired by \cite{chien2005using}) defines constraints and priorities to determine which goal to select among the set of all goals. Furthermore, domain-specific information metrics (e.g., distance traveled and time to perform cost estimation) can aid in goal selection \citep{johnson2016goal}. This work is adapted and generalized to some domains using cost-benefit analysis \citep{kondrakunta2017Implementation,kondrakunta-cox2021}. Trainable-ARTUE \cite{powell2011active} also presents goal selection with expert-based interactive learning. If the system selects a wrong goal, it accepts a penalty.

While the methods presented above take advantage of implementing various goal operations, none of them explicitly shed light on the agent's performance when there is a possibility for interaction of multiple goal operations. This paper addresses the issue by developing a method to prioritize one goal operation over another given the situation in a marine survey domain. The uncertainty in this domain arises from the agent's limited communication when underwater, the unpredictable currents, or the fish attacks. The next section concludes the paper and presents future research.

\section{Conclusions and Future Research}
\label{Conclusion}

In this paper, we explored the structure of decision process when multiple-goal operations co-occur. This paper examines such problems while establishing a communication with the control architecture of the agent. For exploration, the agent performs three search strategies. Initially, we implemented and evaluated each search strategy independently. Subsequently, we introduce goal operations to improve agent performance in the presence of unexpected events. A comparison of agent performance with and without goal operations shows that the agent with goal operations is more robust. To further improve performance, the agent must decide between two goal operations (i.e., goal selection and goal formulation) or go into a safe control mode. The paper develops a response algorithm that aids the agent to make such decisions and maintain its performance in the dynamic world. We compare the performance of this new agent (ASGO) to two other agents (i.e., Select-1st and Formulate-1st). To support our claims of robustness and generality we introduced two new anomalies (i.e., flow, and blockade). The data support our claims that ASGO outperforms others in a dynamic environment.

We intend to extend this research to include several other goal operations (e.g., goal change and goal delegation) \citep{kondrakunta2021grw} in the future. An agent performs goal change operation when it is not able to achieve its current goal due to changes in environment. Goal delegation is useful in a multiagent scenario, for an agent to pass its goals to a different agent. Specifically, goal change is useful when there is variability in the receiver radius. As mentioned in earlier sections, the difference in the water column's stratification can result in variation of the receiver range. From this information, goal change can help the agent decide the amount of resources it needs to spend on the search, such as a sparse search with a high receiver radius or a denser search with a lower radius. Goal delegation can be particularly useful when the agent drifts off the survey region due to flow conditions, or in case of physical damage. It is smart for the agent to delegate its goals to a different agent to complete the mission. Apart from goal operations, we can also integrate the real world parameters. For example, flow prediction parameter for the path planning agent.

\begin{acknowledgements} 
\noindent
The National Science Foundation supported this research under grants 1849131, 1848945, S\&AS-1849137, and also by the Office of Naval Research under grant N00014-18-1-2009. We thank the anonymous reviewers for their comments and suggestions. 
\end{acknowledgements} 

\vspace{-0.25in}

{\parindent -10pt\leftskip 10pt\noindent
\bibliographystyle{cogsysapa}
\bibliography{acs}

}


\end{document}